%% file: parseEmnlp.tex
\documentclass[11pt,a4paper]{article}
\usepackage{acl2015}
\usepackage{times}
\usepackage{url}
\usepackage{latexsym}
\usepackage{multirow}
\usepackage{amssymb}
\usepackage{amsthm}
\usepackage{color}
\usepackage{amsmath}
\usepackage{verbatim}
\usepackage{tikz-qtree}
\usepackage{tikz}
\usetikzlibrary{shapes.geometric}
\usetikzlibrary{shapes.arrows}
\usetikzlibrary{patterns}
\usetikzlibrary{chains}
\usetikzlibrary{calc}

\newcommand{\wfwd}{\stackrel{\rightarrow}{\mathbf{w}}}
\newcommand{\wrev}{\stackrel{\leftarrow}{\mathbf{w}}}

\newcommand{\ignore}[1]{}

\newcommand{\nascomment}[1]{\textcolor{blue}{\textbf{[#1 --\textsc{nas}]}}}

\newsavebox{\one}
\newsavebox{\two}
\newsavebox{\three}
\newsavebox{\four}
\newsavebox{\five}

\title{Improved Transition-Based Parsing \\ by Modeling Characters instead of Words with LSTMs}
\author{Miguel Ballesteros$^{\diamondsuit\spadesuit}$ ~ Chris Dyer$^{\clubsuit\spadesuit}$ ~ Noah A. Smith$^{\heartsuit}$\\
$^\diamondsuit$NLP Group, Pompeu Fabra University, Barcelona, Spain \\
$^\spadesuit$School of Computer Science, Carnegie Mellon University, Pittsburgh, PA, USA \\
$^\clubsuit$Marianas Labs, Pittsburgh, PA, USA \\
$^{\heartsuit}$Computer Science \& Engineering, University of Washington, Seattle, WA, USA\\
{ \sf miguel.ballesteros@upf.edu, chris@marianaslabs.com, nasmith@cs.washington.edu}
}
\date{}

\begin{document}
\maketitle
\begin{abstract}
We present extensions to a continuous-state dependency parsing method that makes it applicable to morphologically rich languages.  Starting with a high-performance transition-based parser that uses long short-term memory (LSTM) recurrent neural networks to learn representations of the parser state, we replace lookup-based word representations with representations constructed from the orthographic representations of the words, also using LSTMs. This allows statistical sharing across word forms that are similar on the surface. Experiments for morphologically rich languages show that the parsing model benefits from incorporating the character-based encodings of words.
\end{abstract}

\section{Introduction}

At the heart of natural language parsing is the challenge of representing the ``state'' of an algorithm---what parts of a parse have been built and what parts of the input string are not yet accounted for---as it incrementally constructs a parse. Traditional approaches rely on independence assumptions, decomposition of scoring functions, and/or greedy approximations to keep this space manageable.
\textbf{Continuous-state} parsers have been proposed, in which the state is embedded as a vector \cite{titov-henderson-2007,stenetorp:2013,chen:2014,lstmacl15,yue:2015,weiss:2015}. Dyer et al.~reported state-of-the-art performance on English and Chinese benchmarks using a transition-based parser whose continuous-state embeddings were constructed using LSTM recurrent neural networks (RNNs) whose parameters were estimated to maximize the probability of a gold-standard sequence of parse actions.

%This paper presents two  extensions to that approach that are critical for application to a larger set of natural languages in which projective parse trees are inadequate \cite{mcdonald05acl} and in which .

The primary contribution made in this work is to take the idea of continuous-state parsing a step further by making the word embeddings that are used to construct the parse state sensitive to the morphology of the words.\footnote{Software for replicating the experiments is available from \url{https://github.com/clab/lstm-parser}.} Since it it is well known that a word's form often provides strong evidence regarding its grammatical role in morphologically rich languages \cite[\emph{inter alia}]{W13-4907}, this has promise to improve accuracy and statistical efficiency relative to traditional approaches that treat each word type as opaque and independently modeled. In the traditional parameterization, words with similar grammatical roles will only be embedded near each other if they are observed in similar contexts with sufficient frequency. Our approach reparameterizes word embeddings using the same RNN machinery used in the parser:  a word's vector is calculated based on the sequence of orthographic symbols representing it (\S\ref{sec:words}).

Although our model is provided no supervision in the form of explicit morphological annotation, we find that it gives a large performance increase when parsing morphologically rich languages in the SPMRL datasets \cite{seddah2013overview,seddah2014introducing}, especially in agglutinative languages and the ones that present extensive case systems (\S\ref{sec:experiments}).  In languages that show little morphology, performance remains good, showing that the RNN composition strategy is capable of capturing both morphological regularities and arbitrariness in the sense of \newcite{saussure-16}.
 Finally, a particularly noteworthy result is that we find that character-based word embeddings in some cases obviate explicit POS information, which is usually found to be indispensable for accurate parsing.%  Taken together, these findings suggest that such representations are able to capture important morphosyntactic information from word strings. Runtime is minimally impacted, and the parser is quite fast.

A secondary contribution of this work is to show that the continuous-state parser of \newcite{lstmacl15} can learn to generate nonprojective trees. We do this by augmenting its transition operations with a \textsc{swap} operation \cite{nivre2009non} (\S\ref{sec:swap}), enabling the parser to produce nonprojective dependencies which are often found in morphologically rich languages. %In this way, the parser is expected to parse nonprojective trees in linear time, and it is comparable with state-of-the-art parsers that are also capable of parsing nonprojective trees \cite{martins:2010,bohnet2012transition}.

\section{An LSTM Dependency Parser}

We begin by reviewing the parsing approach of \newcite{lstmacl15} on which our work is based.

Like most transition-based parsers, Dyer et al.'s parser can be understood as the sequential manipulation of three data structures:  a buffer $B$ initialized with the sequence of words to be parsed,  a stack $S$ containing partially-built parses, and a list $A$ of actions previously taken by the parser. In particular, the parser implements the arc-standard parsing algorithm \cite{nivre2004}. 

At each time step $t$, a transition action is applied that alters these data structures by pushing or popping words from the stack and the buffer; the operations are listed in Figure~\ref{fig:parser}.

Along with the discrete transitions above, the parser calculates a vector representation of the states of $B$, $S$, and $A$; at time step $t$ these are denoted by $\mathbf{b}_t$, $\mathbf{s}_t$, and $\mathbf{a}_t$, respectively.  The total parser state at $t$ is given by
\begin{equation}
\mathbf{p}_t = \max\left\{ \boldsymbol{0}, \mathbf{W}[\mathbf{s}_t; \mathbf{b}_t; \mathbf{a}_t] + \mathbf{d}\right\}
\end{equation}
where the matrix $\mathbf{W}$ and the vector $\mathbf{d}$ are learned parameters.
This continuous-state representation $\mathbf{p}_t$ is used to decide which operation to apply next, updating $B$, $S$, and $A$ (Figure~\ref{fig:parser}).

We elaborate on the design of $\mathbf{b}_t$, $\mathbf{s}_t$, and $\mathbf{a}_t$ using RNNs in \S\ref{sec:stacklstm}, on the representation of partial parses in $S$ in \S\ref{sec:composition},
and on the parser's decision mechanism in \S\ref{sec:decision}. We discuss the inclusion of \textsc{swap} in \S\ref{sec:swap}.

\begin{figure*}
\centering
\begin{tabular}{cc|l|cc|c}
\textbf{Stack}$_t$ & \textbf{Buffer}$_t$ & \textbf{Action} & \textbf{Stack}$_{t+1}$ & \textbf{Buffer}$_{t+1}$ & \textbf{Dependency} \\
\hline
$(\mathbf{u},u),(\mathbf{v},v),S$ & $B$  &$\textsc{reduce-right}(r)$ & $(g_r(\mathbf{u},\mathbf{v}),u),S$ & $B$ & $u \stackrel{\scriptsize{r}}{\rightarrow} v$ \\
$(\mathbf{u},u),(\mathbf{v},v), S$ & $B$ & $\textsc{reduce-left}(r)$ & $(g_r(\mathbf{v},\mathbf{u}),v), S$ & $B$ & $u \stackrel{\scriptsize{r}}{\leftarrow} v$ \\
$S$ & $(\mathbf{u},u),B$ & \textsc{shift} & $(\mathbf{u},u),S$ & $B$ & --- \\ \hline
$(\mathbf{u},u),(\mathbf{v},v),S$ & $B$ & \textsc{swap} & $(\mathbf{u},u),S$ & $(\mathbf{v},v),B$ & ---
\end{tabular}
\caption{\label{fig:parser}Parser transitions indicating the action applied to the stack and buffer and the resulting stack and buffer states. Bold symbols indicate (learned) embeddings of words and relations, script symbols indicate the corresponding words and relations.  \newcite{lstmacl15} used the \textsc{shift} and \textsc{reduce} operations in their continuous-state parser; we add \textsc{swap}.}
\end{figure*}

\subsection{Stack LSTMs} \label{sec:stacklstm}

 RNNs are functions that read a sequence of vectors incrementally;
at time step $t$ the vector $\mathbf{x}_t$ is read in and the hidden state $\mathbf{h}_t$ computed using $\mathbf{x}_t$ and the previous hidden state $\mathbf{h}_{t-1}$.  In principle, this allows retaining information from time steps in the distant past, but the nonlinear ``squashing'' functions applied in the calcluation of each $\mathbf{h}_t$ result in a decay of the error signal used in training with backpropagation. LSTMs are a variant of RNNs designed to cope with this ``vanishing gradient'' problem using an extra memory ``cell'' \cite{hochreiter:1997,graves:2013}. 

Past work explains the computation within an LSTM through the metaphors of deciding how much of the current input to pass into memory ($\mathbf{i}_t$) or forget ($\mathbf{f}_t$).   We refer interested readers to the original papers and present only the recursive equations updating the memory cell $\mathbf{c}_t$ and hidden state $\mathbf{h}_{t}$ given $\mathbf{x}_t$, the previous hidden state $\mathbf{h}_{t-1}$, and the memory cell $\mathbf{c}_{t-1}$:
\begin{align*}
\mathbf{i}_t &= \sigma(\mathbf{W}_{ix}\mathbf{x}_t + \mathbf{W}_{ih}\mathbf{h}_{t-1} + \mathbf{W}_{ic}\mathbf{c}_{t-1} + \mathbf{b}_i) \\
\mathbf{f}_t &= \mathbf{1}-\mathbf{i}_t \\
\mathbf{c}_t &= \mathbf{f}_t \odot \mathbf{c}_{t-1} + \\
& {}\ \ \qquad \mathbf{i}_t \odot \tanh(\mathbf{W}_{cx}\mathbf{x}_t +  \mathbf{W}_{ch}\mathbf{h}_{t-1} + \mathbf{b}_c) \nonumber \\
\mathbf{o}_t &= \sigma(\mathbf{W}_{ox}\mathbf{x}_t + \mathbf{W}_{oh}\mathbf{h}_{t-1} + \mathbf{W}_{oc}\mathbf{c}_{t} + \mathbf{b}_o) \\
\mathbf{h}_t &= \mathbf{o}_t \odot \tanh(\mathbf{c}_t),
\end{align*}
where $\sigma$ is the component-wise logistic sigmoid function and $\odot$ is the component-wise (Hadamard) product.  Parameters are all represented using $\mathbf{W}$ and $\mathbf{b}$. This formulation differs slightly from the classic LSTM formulation in that it makes use of ``peephole connections'' \cite{peeps} and defines the forget gate so that it sums with the input gate to $\mathbf{1}$ \cite{greff:2015}. To improve the representational capacity of LSTMs (and RNNs generally), they can be stacked in ``layers.'' In these architectures, the input LSTM at higher layers at time $t$ is the value of $\mathbf{h}_t$ computed by the lower layer (and $\mathbf{x}_t$ is the input at the lowest layer).
%\nascomment{the ACL paper mentions $\mathbf{y}_t$ but we don't use it so I'm leaving it out ... maybe that's wrong?}
% Finally, output is produced at each time step from the $\mathbf{h}_t$ value at the top layer:
% \begin{align*}
% \mathbf{y}_t &= g(\mathbf{h}_t),
% \end{align*}
% where $g$ is an arbitrary differentiable function. 

The \textbf{stack LSTM} augments the left-to-right sequential model of the conventional LSTM with a stack pointer.  As in the LSTM, new inputs are added in the right-most position, but the stack pointer indicates which LSTM cell provides $\mathbf{c}_{t-1}$ and $\mathbf{h}_{t-1}$ for the computation of the next iterate.  Further, the stack LSTM provides a \textsf{pop} operation that moves the stack pointer to the previous element.  Hence each of the parser data structures ($B$, $S$, and $A$) is implemented with its own stack LSTM, each with its own parameters.  The values of $\mathbf{b}_t$, $\mathbf{s}_t$, and $\mathbf{a}_t$ are the $\mathbf{h}_t$ vectors from their respective stack LSTMs.

\subsection{Composition Functions} \label{sec:composition}

Whenever a \textsc{reduce} operation is selected, two tree fragments are popped off of $S$ and combined to form a new tree fragment, which is then popped back onto $S$ (see Figure~\ref{fig:parser}).  This tree must be embedded as an input vector $\mathbf{x}_t$.

To do this, \newcite{lstmacl15} use a recursive neural network $g_r$ (for relation $r$) that composes the representations of the two subtrees popped from $S$ (we denote these by $\mathbf{u}$ and $\mathbf{v}$), resulting in a new vector $g_r(\mathbf{u}, \mathbf{v})$ or $g_r(\mathbf{v}, \mathbf{u})$, depending on the direction of attachment.  The resulting vector embeds the tree fragment in the same space as the words and other tree fragments.  This kind of composition was thoroughly explored in prior work  \cite{socher:2011,socher:2013,hermann:2013,socher:2013b}; for details, see \newcite{lstmacl15}.

% A particular challenge here is that a syntactic head may, in general, have an arbitrary number of dependents. To simplify the parameterization of our composition function, we combine head-modifier pairs one at a time, building up more complicated structures in the order they are ``reduced'' in the parser, as illustrated in Figure~\ref{fig:composition}. Each node in this expanded syntactic tree has a value computed as a function of its three arguments: the syntactic head ($\mathbf{h}$), the dependent ($\mathbf{d}$), and the syntactic relation being satisfied ($\mathbf{r}$). We define this  by concatenating the vector embeddings of the head, dependent and relation, applying a linear operator and a component-wise nonlinearity as follows:
% \begin{align*}
% \mathbf{c} = \tanh\left(\mathbf{U}[\mathbf{h}; \mathbf{d}; \mathbf{r}] + \mathbf{e} \right).
% \end{align*}
% For the relation vector, we use an embedding of the parser action that was applied to construct the relation (i.e., the syntactic relation paired with the direction of attachment).
% \begin{figure}[h]
% \begin{center}
% \includegraphics[scale=0.75]{composition.pdf}
% \vspace{-0.7cm}
% \end{center}
% \caption{The representation of a dependency subtree (above) is computed by recursively applying composition functions to $\langle \textrm{head}, \textrm{modifier}, \textrm{relation} \rangle$ triples. In the case of multiple dependents of a single head, the recursive branching order is imposed by the order of the parser's reduce operations (below).}
% \label{fig:composition}
% \end{figure}

\subsection{Predicting Parser Decisions} \label{sec:decision}

The parser uses a probabilistic model of parser decisions at each time step $t$.  Letting $\mathcal{A}(S, B)$ denote the set of allowed transitions given the stack $S$ and buffer $S$ (i.e., those where preconditions are met; see Figure~\ref{fig:parser}), the probability of action $z \in \mathcal{A}(S, B)$ defined using a log-linear distribution:
\begin{equation}
p(z \mid \mathbf{p}_t) = \frac{\exp \left( \mathbf{g}_{z}^{\top} \mathbf{p}_t + q_{z} \right)}{\sum_{z' \in \mathcal{A}(S,B)} \exp \left( \mathbf{g}_{z'}^{\top} \mathbf{p}_t + q_{z'} \right)} 
\end{equation}
(where $\mathbf{g}_z$ and $q_z$ are parameters associated with each action type $z$).

Parsing proceeds by always choosing the most probable action from $\mathcal{A}(S, B)$.  The probabilistic definition allows parameter estimation for all of the parameters ($\mathbf{W}_\ast$, $\mathbf{b}_\ast$ in all three stack LSTMs, as well as $\mathbf{W}$, $\mathbf{d}$, $\mathbf{g}_\ast$, and $q_\ast$)
 by maximizing the conditional likelihood of each correct parser decisions given the state. 

\subsection{Adding the \textsc{swap} Operation} \label{sec:swap}
\newcite{lstmacl15}'s parser implemented the most basic version of the arc-standard algorithm, which is capable of producing only projective parse trees. In order to deal with nonprojective trees, we also add the \textsc{swap} operation which allows nonprojective trees to be produced.

The \textsc{swap} operation, first introduced by \newcite{nivre2009non}, allows a transition-based parser to produce nonprojective trees. 
Here, the inclusion of the \textsc{swap} operation requires breaking the linearity of the stack by removing tokens that are not at the top of the stack. This is easily handled with the stack LSTM. Figure \ref{fig:parser} shows how the parser is capable of moving words  from the stack ($S$) to the buffer ($B$), breaking the linear order of words.
Since a node that is swapped may have already been assigned as the head of a dependent, the buffer ($B$) can now also contain tree fragments.

\section{Word Representations} \label{sec:words}

%\nascomment{didn't do much here.  need to give more detail than we have currently.  also suggest fresh notation that doesn't clash at all with what we have above.  reuse the LSTM/RNN nomenclature as much as possible!}

The main contribution of this paper is to change the word representations. In this section, we present the standard word embeddings as in \newcite{lstmacl15}, and the improvements we made generating word embeddings designed to capture morphology based on orthographic strings.

\subsection{Baseline: Standard Word Embeddings}
\label{sec:baselinewords}

%\nascomment{explain how word representations work in the previous paper first, and why this is bad. some things to emphasize:  we are not talking about pretraining or general word embedding learning here; all word embeddings are learned within the parser in this paper (using pretraining is interesting for future work, though, and we should say so)}
%\miguelcomment{I added what you see below. Chris, can you check it? I took it from ACL and rephrase it, but we need to differentiate here, since we do not have the LM model anymore...}

\begin{figure}[h]
%\vspace{-0.42cm}
\includegraphics[scale=0.64]{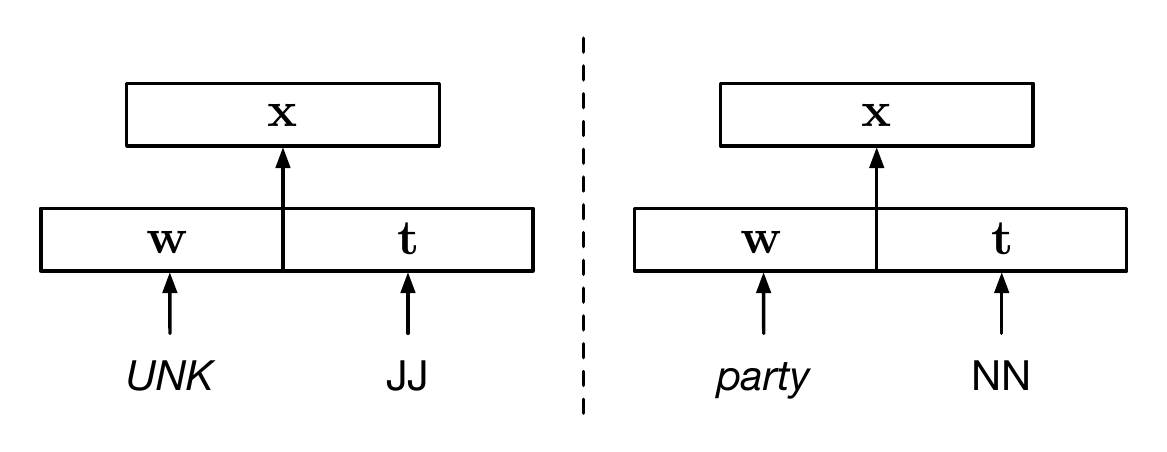}
%\vspace{-2.8cm}
\caption{Baseline model word embeddings for an in-vocabulary word that is tagged with POS tag NN (right) and an out-of-vocabulary word with POS tag JJ (left).% \miguelcomment{ I know it does not print good- Chris: can you take the graffle thing and remove the LM? is this quick}
\label{fig:tokens}}
\end{figure}

Dyer et al.'s parser generates a word representation for each input token by concatenating two vectors: a \ignore{\nascomment{clarify here:  one hot (which is what I initially thought of but seems wrong), or learned-as-parameters?}} vector representation for each word type ($\mathbf{w}$) and a representation ($\mathbf{t}$) of the POS tag of the token (if it is used), provided as auxiliary input to the parser.\footnote{\newcite{lstmacl15}, included a third input representation learned from a neural language model ($\tilde{\mathbf{w}}_{\textrm{LM}}$). We do not include these pretrained representations in our experiments, focusing instead on character-based representations.} A linear map ($\mathbf{V}$) is applied to the resulting vector and passed through a component-wise ReLU:
\begin{align*}
\mathbf{x} = \max\left\{ \mathbf{0}, \mathbf{V}[\mathbf{w}; \mathbf{t}]  + \mathbf{b} \right\} 
\end{align*}
For out-of-vocabulary words, the parser uses an ``UNK" token that is handled as a separate word during parsing time. This mapping can be shown schematically as in Figure~\ref{fig:tokens}.

\subsection{Character-Based Embeddings of Words}
\label{charbased}

\begin{figure}[h]
\centering
\includegraphics[scale=0.64]{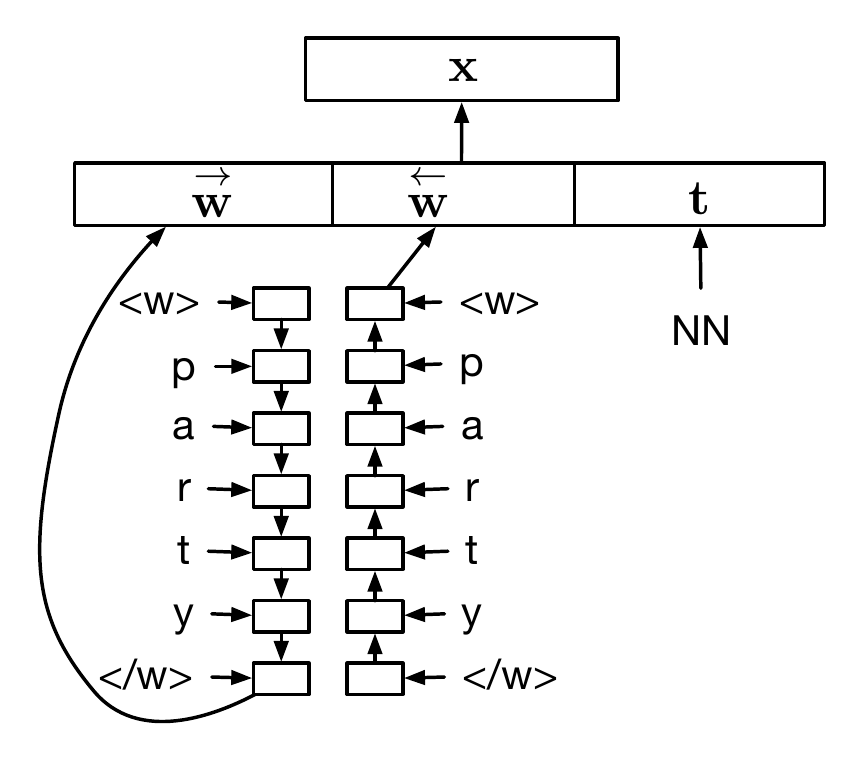}
\caption{Character-based word embedding of the word \emph{party}. This representation is used for both in-vocabulary and out-of-vocabulary words. %\miguelcomment{ I know it does not print good- Chris: could you take the graffle thing and do something like this?}
\label{fig:chartokens}}
\end{figure}

Following \newcite{ling:2015}, we compute character-based 
 continuous-space vector embeddings of words using bidirectional LSTMs \cite{journals/nn/GravesS05}. When the parser initiates the learning process and populates the buffer with all the words from the sentence, it reads the words character by character from left to right and computes a continuous-space vector embedding the character sequence, which is the $\mathbf{h}$ vector of the LSTM; we denote it by $\wfwd$. The same process is also applied in reverse (albeit with different parameters), computing a similar continuous-space vector embedding starting from the last character and finishing at the first ($\wrev$); again each character is represented with an LSTM cell. After that, we concatenate these vectors and a (learned) representation of their tag to produce the representation $\mathbf{w}$. As in \S\ref{sec:baselinewords}, a linear map ($\mathbf{V}$) is applied and passed through a component-wise ReLU. 
\begin{align*}
\mathbf{x} &= \max\left\{ \mathbf{0}, \mathbf{V}[\wfwd; \wrev; \mathbf{t}]  + \mathbf{b} \right\}
\end{align*}
This process is shown schematically in Figure~\ref{fig:chartokens}.

Note that under this representation, out-of-vocabulary words are treated as bidirectional LSTM encodings and thus they will be ``close'' to other words that the parser has seen during training, ideally close to their more frequent, syntactically similar morphological relatives. We conjecture that this will give a  clear advantage over a single ``UNK'' token for all the words that the parser does not see during training, as done by \newcite{lstmacl15} and other parsers without additional resources.
In \S\ref{sec:experiments} we confirm this hypothesis.

\ignore{\begin{figure*}[!ht]
\centering
\includegraphics[scale=0.45]{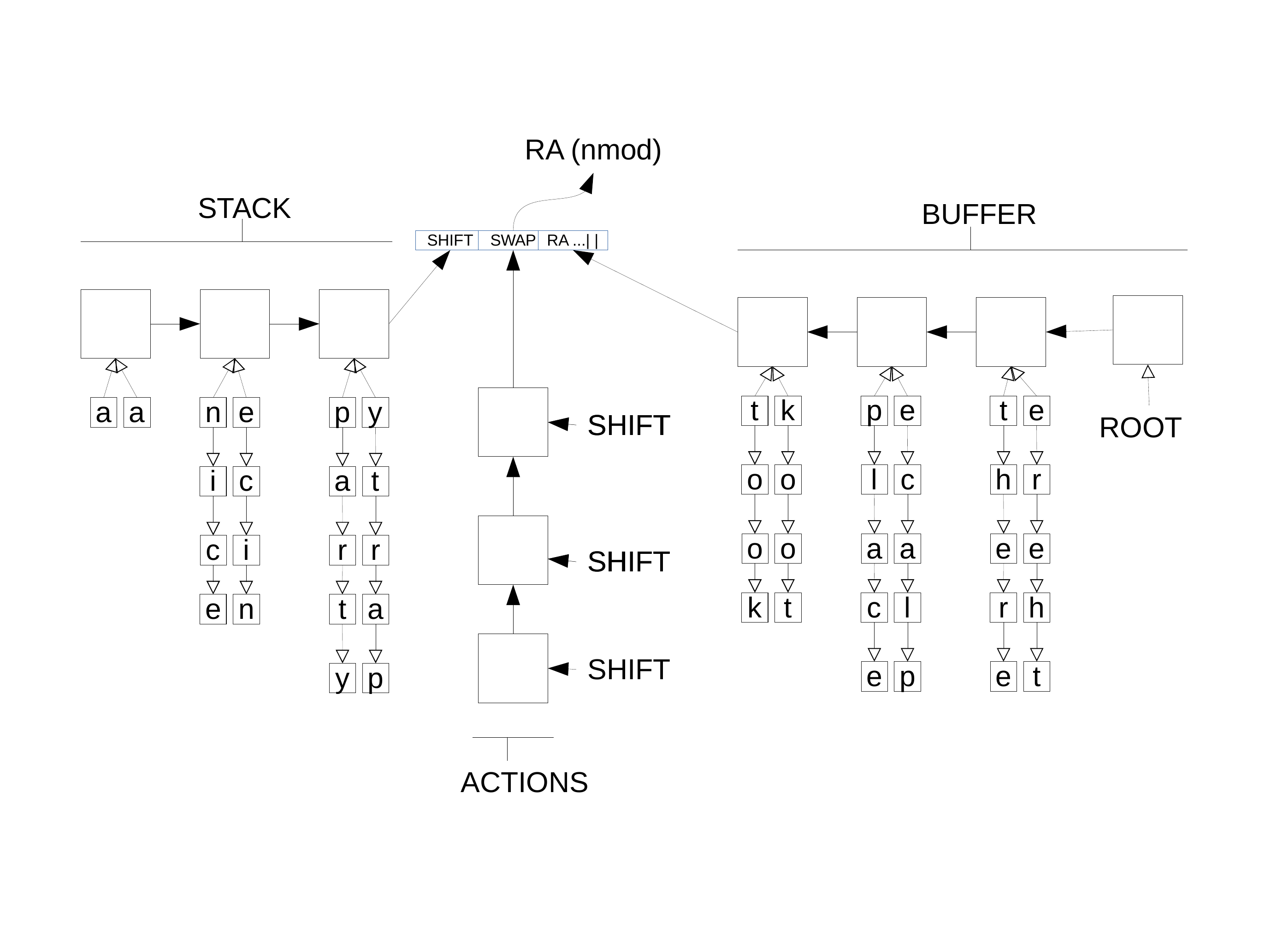}
\vspace{-1.7cm}
\caption{Parser state computation encountered while parsing the sentence ``\emph{a nice party took place there}.'' \emph{STACK} is the stack of partially constructed syntactic subtrees and its LSTM encoding including the LSTM encoding of characters for each word; \emph{BUFFER} is the buffer of words remaining to be processed and its LSTM encoding including the LSTM encoding of characters for each word; and \emph{ACTIONS} is the stack representing the history of actions taken by the parser. These are linearly transformed, passed through a ReLU nonlinearity to produce the parser state embedding.}
\label{transition}
\end{figure*}}

\section{Experiments}
\label{sec:experiments}

%\nascomment{did not do anything here.  I suggest reframing this to focus first on multilingual experiments (include E and C later, for ``completeness'').  the new intro highlights what I think are the key findings.  Ideally, we'd show the benefits of swap and of character level embeddings separately as well as together.  I think separating out the no-POS versions in a different subsection is the right move.  the punchline, I think, is that these representation learners are getting word represenatations that capture morphosyntactic information.}

We applied our parsing model and several variations of it to several parsing tasks and report results below.

\subsection{Data}

In order to find out whether the character-based representations are capable of learning the morphology of words, we applied the parser to morphologically rich languages specifically the treebanks of the SPMRL shared task \cite{seddah2013overview,seddah2014introducing}: Arabic \cite{maamouri04arabic}, Basque \cite{basque:03}, French \cite{french:03}, German \cite{tiger:dep:2012}, Hebrew \cite{hebrew:01}, Hungarian \cite{vincze2010hungarian}, Korean \cite{Choi:Kaistdep:Spmrl}, Polish \cite{polish:10} and Swedish \cite{swedish:06}. For all the corpora of the SPMRL Shared Task we used predicted POS tags as provided by the shared task organizers.\footnote{The POS tags were calculated with the MarMot tagger \cite{mueller2013} by the best performing system of the SPMRL Shared Task \cite{bjorkelund-EtAl:2013:SPMRL}. Arabic: 97.38. Basque: 97.02. French: 97.61. German: 98.10. Hebrew: 97.09. Hungarian: 98.72. Korean: 94.03. Polish: 98.12. Swedish: 97.27.}  For these datasets, evaluation is calculated using \url{eval07.pl}, which includes punctuation.

We also experimented with the Turkish dependency treebank\footnote{Since the Turkish dependency treebank does not have a development set, we extracted the last 150 sentences from the 4996 sentences of the training set as a development set.} \cite{oflazer2003building} of the CoNLL-X Shared Task \cite{buchholz06}.  We used gold POS tags, as is common with the CoNLL-X data sets.

To put our results in context with the most recent neural network transition-based parsers, we run the parser in the same Chinese and English setups as \newcite{chen:2014} and \newcite{lstmacl15}.
For Chinese, we use the Penn Chinese Treebank 5.1 (CTB5) following Zhang and Clark \shortcite{zhang08},\footnote{Training: 001--815, 1001--1136. Development: 886--931, 1148--1151. Test: 816--885, 1137--1147.} with gold POS tags.
For English, we used the Stanford Dependency (SD) representation of the Penn Treebank\footnote{Training: 02--21. Development: 22. Test: 23.} \cite{marcus-93,Marneffe06generatingtyped}.\footnote{The POS tags are predicted by using the Stanford Tagger \cite{Toutanova:2003:FPT:1073445.1073478} with an accuracy of 97.3\%.}. % for which we run the MarMoT tagger \cite{mueller2013}, with an accuracy of 93.9\% in the development set, with ten-way jackknifing of the training data. This treebank is fully projective.
Results for Turkish, Chinese, and English are calculated using the CoNLL-X \url{eval.pl} script, which ignores punctuation symbols.

\subsection{Experimental Configurations}

In order to isolate the improvements provided by the LSTM encodings of characters, we run the stack LSTM parser in the following configurations:
\begin{itemize}
\item \textbf{Words}:  words only, as in \S\ref{sec:baselinewords} (but without POS tags)
\item \textbf{Chars}:  character-based representations of words with bidirectional LSTMs, as in \S\ref{charbased} (but without POS tags)
\item \textbf{Words + POS}:  words and POS tags (\S\ref{sec:baselinewords})
\item \textbf{Chars + POS}:  character-based representations of words with bidirectional LSTMs plus POS tags (\S\ref{charbased})
\end{itemize}

None of the experimental configurations include pretrained word-embeddings or any additional data resources. All experiments include the \textsc{swap} transition, meaning that nonprojective trees can be produced in any language. % Note that we are not using explicit morphological feature, only POS tags and word forms since we want to isolate the effect of the character-based representations.

%\nascomment{somewhere we have to say how many cells we used in each LSTM.} I think it is shown above?

\paragraph{Dimensionality.}
The full version of our parsing model sets dimensionalities as follows.  LSTM hidden states are of size 100, and we use two layers of LSTMs for each stack. Embeddings of the parser actions used in the composition functions have 20 dimensions, and the output embedding size is 20 dimensions. The learned word representations embeddings have 32 dimensions when used, while the character-based representations have 100 dimensions, when used. Part of speech embeddings have 12 dimensions. These dimensionalities were chosen after running several tests with different values, but a more careful selection of these values would probably further improve results.

\subsection{Training Procedure}
Parameters are initialized randomly---refer to \newcite{lstmacl15} for specifics---and optimized using stochastic gradient descent (without minibatches) using derivatives of the negative log likelihood of the sequence of parsing actions computed using backpropagation. Training is stopped when the learned model's UAS stops improving on the development set, and this model is used to parse the test set. No pretraining of any parameters is done.

\subsection{Results and Discussion}

%\nascomment{I rearranged the tables to make it easier to compare within UAS and LAS.  I also rewrote this section.  Please check carefully.} Miguel: done! thanks, it reads nicely.

\input{new-table}

Tables \ref{devresults} and \ref{testresults} show the results of the parsers for the development sets and the final test sets, respectively.  Most notable are improvements for agglutinative languages---Basque, Hungarian, Korean, and Turkish---both when POS tags are included and when they are not.  Consistently, across all languages, \textbf{Chars} outperforms \textbf{Words}, suggesting that the character-level LSTMs are learning representations that capture similar information to parts of speech.  On average, \textbf{Chars} is on par with \textbf{Words + POS}, and the best average of labeled attachment scores is achieved with \textbf{Chars + POS}. 

%\nascomment{this is not consistent with the tables:}
%In fact, in Arabic, Basque, Korean, and Turkish [Char-based] is better for LAS than [Char-based + POS] and [Words + POS], in the final test sets.

It is common practice to encode morphological information in treebank POS tags; for instance, the Penn Treebank includes 
English number and tense (e.g., \textsc{NNS} is plural noun and \textsc{VBD} is verb in past tense).
 Even if our character-based representations are capable of encoding the same kind of information, existing POS tags suffice for high accuracy.  However, the POS tags in treebanks for morphologically rich languages do not seem to be enough.

Swedish, English, and French use suffixes for the verb tenses and number,\footnote{Tense and number features provide little improvement in a transition-based parser, compared with other features such as case, when the POS tags are included \cite{W13-4907}.} while Hebrew uses prepositional particles rather than grammatical case. \newcite{tsarfaty2006integrated} and \newcite{cohen07} argued that, for Hebrew, determining the correct morphological segmentation is dependent on syntactic context. Our approach sidesteps this step, capturing the same kind of information in the vectors, and learning it from syntactic context.   Even for Chinese, which is not morphologically rich, \textbf{Chars} shows a benefit over \textbf{Words}, perhaps by capturing regularities in syllable structure within words. %\nascomment{check here; I commented out the earlier sentence, which I'm not sure I understood}

% Chinese morphology is mainly represented by syllables that can even stand alone as different words, nonetheless the character-based representations are still useful for Chinese.

\subsubsection{Learned Word Representations}

\begin{figure}[!ht]
%\centering
\includegraphics[scale=.29]{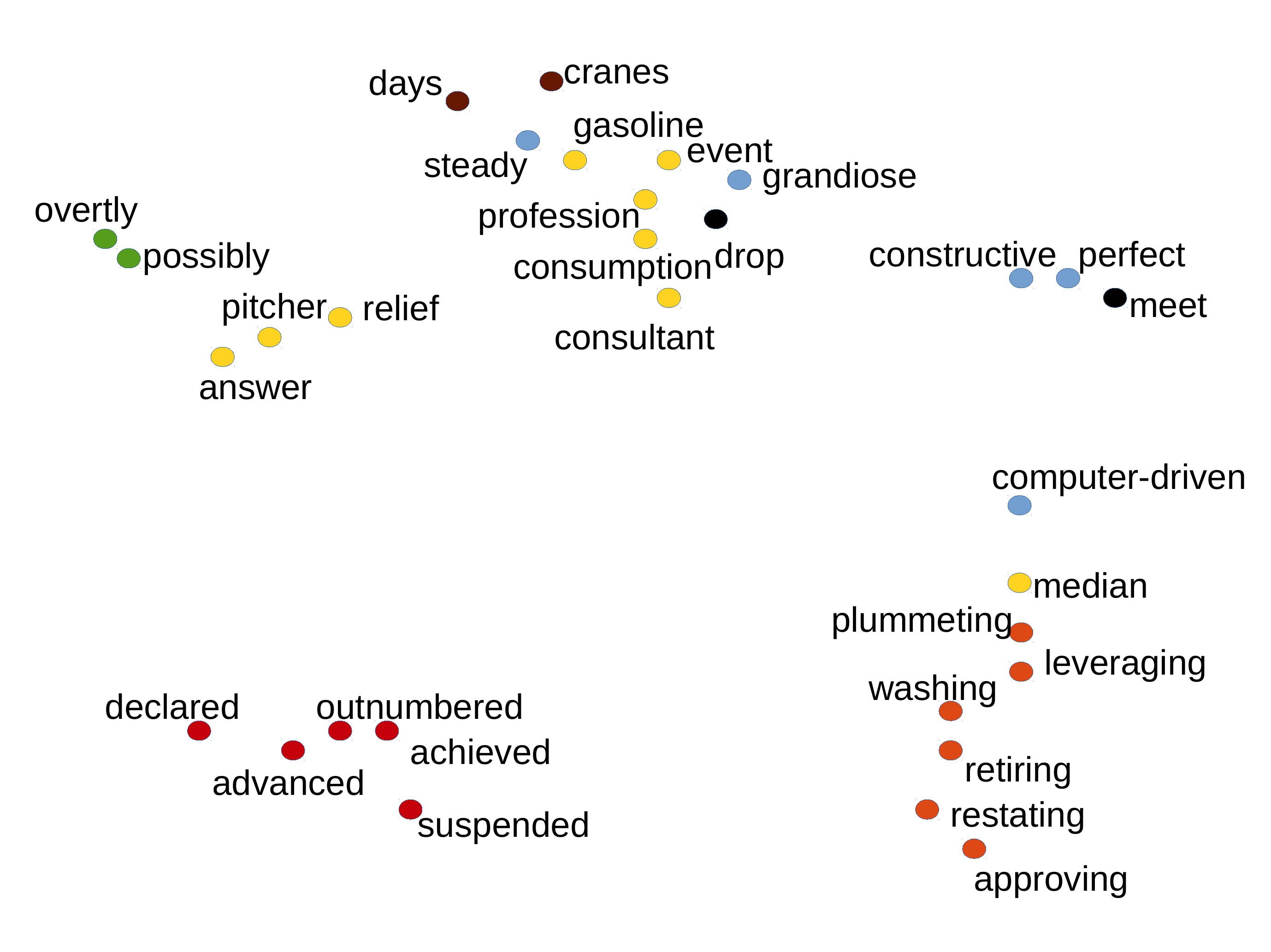}
\caption{Character-based word representations of 30 random words from the English development set (\textbf{Chars}). Dots in red represent past tense verbs; dots in orange represent gerund verbs; dots in black represent present tense verbs; dots in blue represent adjectives; dots in green represent adverbs; dots in yellow represent singular nouns; dots in brown represent plural nouns.  The visualization was produced using t-SNE; see \url{http://lvdmaaten.github.io/tsne/}.}%\cjd{Could we have different colors for the dots based on their most frequent gold POS tags?}}
\label{charbasedvectors}
\end{figure}

%Figure \ref{wordvectors} plots the learned representations for 50 random words in the English development set, for the standard word vectors [Words]. While 
Figure \ref{charbasedvectors} visualizes a sample of the character-based bidirectional LSTMs's learned representations (\textbf{Chars}).   Clear clusters of past tense verbs, gerunds, and other syntactic classes are visible. The colors in the figure represent the most common POS tag for each word.
%\nascomment{could we show more?  some of these words are ambiguous; reviewers might complain about 'complete' and 'drop' ... maybe only show unambiguous ones?  also, can the font be larger?}
%However, the word vectors shown in Figure \ref{wordvectors} are less clear, and thus, as shown in Tables \ref{developmentresults} and \ref{testresults} they consistently tend to provide worse results for the parsing task.

%\begin{figure}[!ht]
%\centering
%\includegraphics[scale=0.35]{wordvectors2.png}
%\caption{Learned representations as standard word vectors of 50 random words from the English development set [Words].}
%\label{wordvectors}
%\end{figure}

\subsubsection{Out-of-Vocabulary Words}

\begin{figure*}
\centering \includegraphics[width=0.9\textwidth]{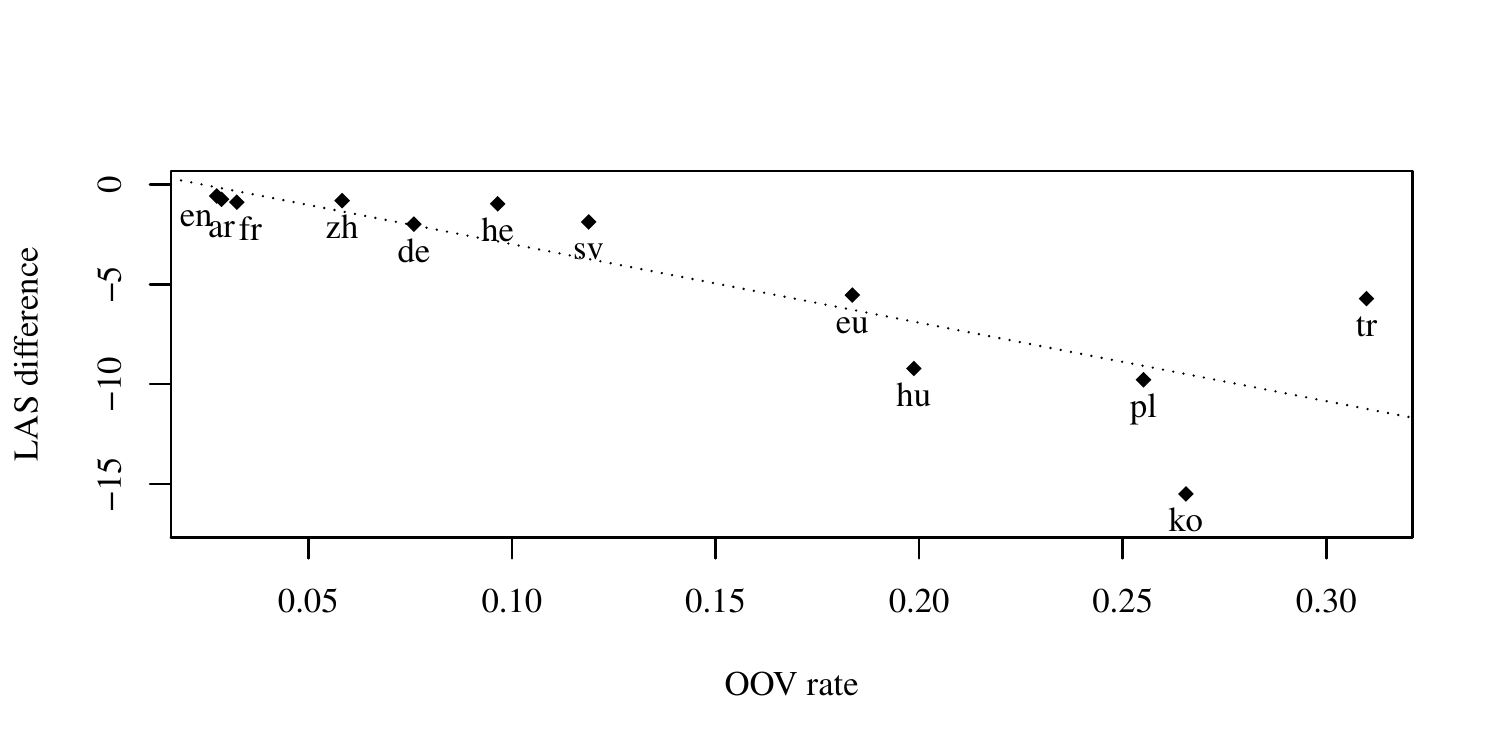}
\caption{On the $x$-axis is the OOV rate in development data, by treebank; on the $y$-axis is the difference in development-set LAS between \textbf{Chars} model as described in \S\ref{charbased} and one in which all OOV words are given a single representation. \label{oovplot}}
\end{figure*}

The character-based representation for words is notably beneficial for out-of-vocabulary (OOV) words.  We tested this specifically by comparing \textbf{Chars} to a model in which all OOVs are replaced by the string ``UNK'' during parsing.
This always has a negative effect on LAS (average $-4.5$ points, $-2.8$ UAS).  Figure~\ref{oovplot} shows how this drop varies with the development OOV rate across treebanks; most extreme is Korean, which drops 15.5 LAS.  A similar, but less pronounced pattern, was observed for models that include POS.

Interestingly, this artificially impoverished model is still consistently better than \textbf{Words} for all languages (e.g., for Korean, by 4 LAS).  This implies that not all of the improvement is due to OOV words; statistical sharing across orthographically close words is beneficial, as well.

%\nascomment{I commented out some details here that don't seem that important}
% Including POS tags, [Char-based + POS] is also better for most languages but, as expected, the improvements are not that high. The highest one is Hungarian with 5.14 LAS points of improvement (71.02 vs 76.16) in the development set. Basque drops 3.49 LAS points (2,541 OOVs out of 13,840 tokens). In other languages, such as French, there is just a small variance (+/-0.02 LAS points).

%More details are shown in the supplementary materials.

\subsubsection{Computational Requirements}

The character-based representations make the parser slower, since they require composing the character-based bidirectional LSTMs for each word of the input sentence; however, at test time these results could be cached. On average, \textbf{Words} parses a sentence in 44 ms, while\textbf{Chars} needs 130 ms.\footnote{We are using a machine with 32 Intel Xeon CPU E5-2650 at 2.00GHz; the parser runs on a single core.} Training time is affected by the same cons\-tant, needing some hours to have a competitive model.
In terms of memory, \textbf{Words} requires on average 300 MB of main memory for both training and parsing, while \textbf{Chars} requires 450 MB.

%In terms of memory, [Words] requires, in mean, 300 MB of main memory for both training and parsing, while [Char-based] requires 450 MB of main memory for both training and parsing.
%\miguelcomment{The memory is something worth noting, since some parsers need up to 30GB (or even more) for training and parsing. Mention minibatches here? or remove entire subsection?}

\subsubsection{Comparison with State-of-the-Art}

%\miguelcomment{I added a table here, since I don't want to deviate the attention of the reader to state-of-the-art performance... in the other tables... the other one is commented if you think we sould do it otherwise.}

\begin{table*}[!ht]
\small
\centering
\begin{tabular}{|l||r|r|l||r|r|l||r|r|l|}
\hline
%\multicolumn{7}{|c|}{ Shared Task on Parsing Morphologically Rich Languages}   \\ \hline
& \multicolumn{3}{|c||}{\textbf{This Work}} & \multicolumn{3}{|c||}{\textbf{Best Greedy Result}} & \multicolumn{3}{|c|}{\textbf{Best Published Result}}
\\
\hline
 \bf Language & \bf UAS &\bf  LAS & \bf System & \bf UAS &\bf  LAS & \bf System & \bf UAS &\bf  LAS & \bf System
\\
\hline
Arabic  & 86.08 & 83.41 & \textbf{Chars} & 84.57 & 81.90 & B'13 & 88.32 & 86.21 & B+'13
\\
Basque  & 85.22 & 78.61 & \textbf{Chars + POS} & 84.33 & 78.58 & B'13 &89.96 & 85.70 & B+'14
\\
French & 86.15 & 82.03 & \textbf{Words + POS} & 83.35 & 77.98 & B'13 &89.02 & 85.66 & B+'14
\\
German  & 87.33 & 84.62 & \textbf{Words + POS} & 85.38 & 82.75 & B'13 &91.64 & 89.65 & B+'13
\\
Hebrew & 80.68 & 72.70 & \textbf{Words + POS} &  79.89 & 73.01 & B'13 &87.41 & 81.65 & B+'14
\\
Hungarian & 80.92 & 76.34 & \textbf{Chars + POS} & 83.71 & 79.63 & B'13 &89.81 & 86.13 & B+'13
\\
Korean & 88.39 & 86.27 & \textbf{Chars} & 85.72 & 82.06 & B'13 &89.10 & 87.27 & B+'14
\\
Polish & 87.06 & 79.83 & \textbf{Words + POS} & 85.80 & 79.89 & B'13 &91.75 & 87.07 & B+'13
\\
Swedish & 83.43 & 76.40 & \textbf{Words + POS} & 83.20 & 75.82 & B'13 &88.48 & 82.75 & B+'14
\\
\hline
Turkish & 76.32 & 64.34 & \textbf{Chars} & 75.82 & 65.68 & N+'06a &77.55 & n/a & K+'10
\\
\hline
Chinese  & 85.96 & 84.40 &\textbf{Words + POS} & 87.20 & 85.70 & D+'15& 87.20 & 85.70 &  D+'15
\\
English  & 92.57 & 90.31 & \textbf{Words + POS} & 93.10 & 90.90 & D+'15& 94.08 & 92.19 & W+'15
\\
\hline
\end{tabular}
%\caption{Unlabelled attachment scores (UAS) and Labelled attachment scores (LAS) results in the test sets. The column [Ours] show the results of our parser in its best performing configuration for UAS. The Column [Greedy] shows the results for transition-based greedy parsers in these data sets which is the most fair comparison to our results: for English and Chinese, it shows our own best results of \newcite{lstmacl15}'s parser with pretrained word embeddings; for the SPMRL data sets it shows the results of the MaltOptimizer submission to the SPMRL shared task \cite{W13-4907}, and for Turkish, it shows the MaltParser submission to the CoNLL-X Shared Task \cite{nivre06conll}. The Column [Best] shows the best results ever reported in these data sets. For the SPMRL treebanks, it shows the results of \newcite{}; for Turkish \cite{Koo:2010:DDP:1870658.1870783}; for English \cite{} and for Chinese \cite{}.}
\caption{Test-set performance of our best results (according to UAS or LAS, whichever has the larger difference), compared to state-of-the-art greedy transition-based parsers (``Best Greedy Result'') and best results reported (``Best Published Result'').  All of the systems we compare against use explicit morphological features and/or one of the following: pretrained word embeddings, unlabeled data and a combination of parsers; our models do not.
B'13 is \newcite{W13-4907}; N+'06a is \newcite{nivre06conll}; D+'15 is \newcite{lstmacl15};
B+'13 is \newcite{bjorkelund-EtAl:2013:SPMRL}; B+'14 is \newcite{bjorkelund-EtAl:2014:SPMRL-SANCL};
K+'10 is \newcite{Koo:2010:DDP:1870658.1870783};
W+'15 is  \newcite{weiss:2015}.
\label{compsota}}
\end{table*}

Table \ref{compsota} shows a comparison with state-of-the-art parsers.  We include greedy transition-based parsers that, like ours, do not apply a beam search \cite{zhang08} or a dynamic oracle \cite{goldberg2013training}.  For all the SPMRL languages we show the results of \newcite{W13-4907}, who reported results after carrying out a careful automatic morphological feature selection experiment. For Turkish, we show the results of \newcite{nivre06conll} which also carried out a careful manual morphological feature selection.  Our parser outperforms these in most cases.  Since those systems rely on morphological features, we believe that this comparison shows even more that the character-based representations are capturing morphological information, though without explicit morphological features. For English and Chinese, we report \cite{lstmacl15} which is \textbf{Words + POS} but with pretrained word embeddings.

%It is worth noting that it is a \textbf{greedy} transition-based dependency parser and it does not use neither additional resources, such as pretrained word embeddings, nor explicit morphological features. For English, our best result is comparable with the state-of-the-art, 0.5 points worse than the same parser with pretrained word embeddings (93.1) \cite{lstmacl15} and better than \newcite{chen:2014} parser that also used pretrained word embeddings. For Chinese, the result is even more competitive, higher than \newcite{chen:2014} parser (84.00) and close to the best result published in \cite{lstmacl15}.

We also show the best reported results on these datasets.  For the SPMRL data sets, the best performing system of the shared task is either \newcite{bjorkelund-EtAl:2013:SPMRL} or \newcite{bjorkelund-EtAl:2014:SPMRL-SANCL}, which are consistently better than our system for all languages.
Note that the comparison is harsh to our system, which does not use unlabeled data or explicit morphological features nor any combination of different parsers.
For Turkish, we report the results of \newcite{Koo:2010:DDP:1870658.1870783}, which only reported unlabeled attachment scores. For English, we report \cite{weiss:2015} and for Chinese, we report \cite{lstmacl15} which is \textbf{Words + POS} but with pretrained word embeddings.

% turkish best result
%77.55 (Ko10)
%Terry Koo, Alexander M Rush, Michael Collins,
%Tommi Jaakkola, and David Sontag. 2010. Dual
%decomposition for parsing with nonprojective head
%automata. In Proceedings of the 2010 Conference on
%Empirical Methods in Natural Language Processing.
%Association for Computational Linguistics.

\section{Related Work}
\label{sec:related}

Character-based representations have been explored in other NLP tasks; for instance, \newcite{santos2014learning} and \newcite{nerchars} learned character-level neural representations for POS tagging and named entity recognition, getting a large error reduction in both tasks. Our approach is similar to theirs.
Others have used character-based models as features to improve existing models. For instance, \newcite{chrupalanormalizing} used character-based recurrent neural networks to normalize tweets.

\newcite{Botha2014} show that stems, prefixes and suffixes can be used to learn useful word representations but relying on an external morphological analyzer. That is, they learn the morpheme-meaning relationship with an additive model, whereas we do not need a morphological analyzer. Similarly, \newcite{chenjoint} proposed joint learning of character and word embeddings for Chinese, claiming that characters contain rich information.% \nascomment{what is ``internal information''?}.

Methods for joint morphological disambiguation and parsing have been widely explored
\newcite{tsarfaty2006integrated,cohen07,goldberg2008single,goldberg2011joint}.
More recently, \newcite{tacl-bbjn} presented an arc-standard transition-based parser that performs competitively for joint morphological tagging and dependency parsing for richly inflected languages, such as Czech, Finnish, German, Hungarian, and Russian.
Our model seeks to achieve a similar benefit to parsing without explicitly reasoning about the internal structure of words.

\newcite{zhang2013chinese} presented efforts on Chinese parsing with characters showing that Chinese can be parsed at the character level, and that Chinese word segmentation is useful for predicting the correct POS tags \cite{zhang2008joint}.

To the best of our knowledge, previous work has not used character-based embeddings to improve dependency parsers, as done in this paper.

\section{Conclusion}

We have presented several interesting findings.  First, we add new evidence that character-based representations are useful for NLP tasks. In this paper, we demonstrate that they are useful for transition-based dependency parsing, since they are capable of capturing morphological information crucial for analyzing syntax.

The improvements provided by the character-based representations using bidirectional LSTMs are strong for agglutinative languages, such as Basque, Hungarian, Korean, and Turkish, comparing favorably to POS tags as encoded in those languages' currently available treebanks. This outcome is important, since annotating morphological information for a treebank is expensive.  Our finding suggests that the best investment of annotation effort may be in dependencies, leaving morphological features to be learned implicitly from strings.

The character-based representations are also a way of overcoming the out-of-vocabulary problem; without any additional resources, they enable  the parser to substantially improve the performance when OOV rates are high. We expect that, in conjunction with a pretraing regime, or in conjunction with distributional word embeddings, further improvements could be realized.

%Finally, as a parallel outcome of the experiment, \newcite{lstmacl15}'s parser was augmented with the \textsc{swap} operation that makes it capable of producing nonprojective parse trees.

\ignore{
\bigskip
\noindent
{\centering
\fbox{
\begin{minipage}{7cm}
\textbf{Note:} We are submitting supplementary material showing more results about OOV words and the full set of development results.
\end{minipage}
}}}

\section*{Acknowledgments}
MB was supported by the European Commission under the contract numbers FP7-ICT-610411 (project MULTISENSOR) and H2020-RIA-645012 (project KRISTINA).
This research was supported by the U.S.~Army Research Laboratory and the U.S.~Army Research Office
under contract/grant number W911NF-10-1-0533 and NSF IIS-1054319.  This work was completed while NAS was at CMU. Thanks to Joakim Nivre, Bernd Bohnet, Fei Liu and Swabha Swayamdipta for useful comments.

%\nascomment{to do:  clean up bibliography.  be consistent about capitalization, whether we use names or initials, punctuation within initials, names of conferences, whether we use page numbers, etc.}

\bibliographystyle{acl}
\bibliography{biblio,main}

\end{document}

%% file: new-table.tex
\begin{table*}[!ht]
\centering
\small
\begin{tabular}{|l|r|r|r|r|}
\multicolumn{5}{c}{UAS} \\
\hline
\bf \multirow{2}{*}{Language} & \bf \multirow{2}{*}{Words} & \bf \multirow{2}{*}{Chars} & \bf Words & \bf Chars  \\ 
&&&\bf + POS & \bf + POS \\
\hline
Arabic & 86.14 & \bf87.20 & \bf87.44 & 87.07
\\
Basque & 78.42 & \bf84.97 & 83.49 & \bf85.58
\\
French & 84.84 & \bf86.21 & \bf87.00 & 86.33
\\
German & 88.14 & \bf90.94 & 91.16 & \bf91.23
\\
Hebrew & 79.73 & \bf79.92 & \bf81.99 & 80.76
\\
Hungarian & 72.38 & \bf80.16 & 78.47 & \bf80.85
\\
Korean & 78.98 & \bf88.98 & 87.36 & \bf89.14
\\
Polish & 73.29 & \bf85.69 & \bf89.32 & 88.54
\\
Swedish & 73.44 & \bf75.03 & \bf80.02 & 78.85
\\
\hline
Turkish & 71.10 & \bf74.91 & 77.13 & \bf77.96
\\
\hline
Chinese & 79.43 & \bf80.36 & \bf85.98 & 85.81
\\
English & 91.64 & \bf91.98 & \bf92.94 & 92.49
\\
\hline
Average & 79.79 & \bf83.86 & 85.19 & \bf85.38
\\
\hline
\end{tabular} \hspace{0.25in}
\begin{tabular}{|l|r|r|r|r|} 
\multicolumn{5}{c}{LAS} \\
\hline
\bf \multirow{2}{*}{Language} & \bf \multirow{2}{*}{Words} & \bf \multirow{2}{*}{Chars} & \bf Words & \bf Chars  \\ 
&&&\bf + POS & \bf + POS \\
\hline
Arabic & 82.73 & \bf84.34 & \bf84.81 & 84.36
\\
Basque & 67.08 & \bf78.22 & 74.31 & \bf79.52
\\
French & 80.32 & \bf81.70 & \bf82.71 & 81.51
\\
German & 85.36 & \bf88.68 & \bf89.04 & 88.83
\\
Hebrew & 69.42 & \bf70.58 & \bf74.11 & 72.18
\\
Hungarian & 62.14 & \bf75.61 & 69.50 & \bf76.16
\\
Korean & 67.48 & \bf86.80 & 83.80 & \bf86.88
\\
Polish & 65.13 & \bf78.23 & \bf81.84 & 80.97
\\
Swedish & 64.77 & \bf66.74 & \bf72.09 & 69.88
\\
\hline
Turkish & 53.98 & \bf62.91 & 62.30 & \bf62.87
\\
\hline
Chinese & 75.64 & \bf77.06 & \bf84.36 & 84.10
\\
English & 88.60 & \bf89.58 & \bf90.63 & 90.08
\\
\hline
Average & 71.89 & \bf78.37 & 79.13 & \bf79.78
\\
\hline
\end{tabular}
\caption{Unlabeled attachment scores (left) and labeled attachment scores (right) on the \textbf{development} sets (not a standard development set for Turkish). In each table, the first two columns show the results of the parser with word lookup (\textbf{Words}) vs.~character-based (\textbf{Chars}) representations. The last two columns add POS tags.  Boldface shows the better result comparing \textbf{Words} vs.~\textbf{Chars} and comparing \textbf{Words + POS} vs.~\textbf{Chars + POS}. \label{devresults} }
\end{table*}

\begin{table*}[!ht]
\centering
\small
\begin{tabular}{|l|r|r|r|r|}
\multicolumn{5}{c}{UAS} \\
\hline
\bf \multirow{2}{*}{Language} & \bf \multirow{2}{*}{Words} & \bf \multirow{2}{*}{Chars} & \bf Words & \bf Chars  \\ 
&&&\bf + POS & \bf + POS \\
\hline
Arabic & 85.21 & \bf86.08 & 86.05 & \bf86.07
\\
Basque & 77.06 & \bf85.19 & 82.92 & \bf85.22
\\
French & 83.74 & \bf85.34 & \bf86.15 & 85.78
\\
German & 82.75 & \bf86.80 & \bf87.33 & 87.26
\\
Hebrew & 77.62 & \bf79.93 & \bf80.68 & 80.17
\\
Hungarian & 72.78 & \bf80.35 & 78.64 & \bf80.92
\\
Korean & 78.70 & \bf88.39 & 86.85 & \bf88.30
\\
Polish & 72.01 & \bf83.44 & \bf87.06 & 85.97
\\
Swedish & 76.39 & \bf79.18 & \bf83.43 & 83.24
\\
\hline
Turkish & 71.70 & \bf76.32 & 75.32 & \bf76.34
\\
\hline
Chinese & 79.01 & \bf79.94 & \bf85.96 & 85.30
\\
English & 91.16 & \bf91.47 & \bf92.57 & 91.63
\\
\hline
Average & 79.01 & \bf85.36 & 84.41 & \bf84.68
\\
\hline
\end{tabular} \hspace{0.25in}
\begin{tabular}{|l|r|r|r|r|} 
\multicolumn{5}{c}{LAS} \\
\hline
\bf \multirow{2}{*}{Language} & \bf \multirow{2}{*}{Words} & \bf \multirow{2}{*}{Chars} & \bf Words & \bf Chars  \\ 
&&&\bf + POS & \bf + POS \\
\hline
Arabic & 82.05 & \bf83.41 & \bf83.46 & 83.40
\\
Basque & 66.61 & \bf79.09 & 73.56 & \bf78.61
\\
French & 79.22 & \bf80.92 & \bf82.03 & 81.08
\\
German & 79.15 & \bf84.04 & \bf84.62 & 84.49
\\
Hebrew & 68.71 & \bf71.26 & \bf72.70 & 72.26
\\
Hungarian & 61.93 & \bf75.19 & 69.31 & \bf76.34
\\
Korean & 67.50 & \bf86.27 & 83.37 & \bf86.21
\\
Polish & 63.96 & \bf76.84 & \bf79.83 & 78.24
\\
Swedish & 67.69 & \bf71.19 & \bf76.40 & 74.47
\\
\hline
Turkish & 54.55 & \bf64.34 & 61.22 & \bf62.28
\\
\hline
Chinese & 74.79 & \bf76.29 & \bf84.40 & 83.72
\\
English & 88.42 & \bf88.94 & \bf90.31 & 89.44
\\
\hline
Average & 71.22 & \bf78.15 & 78.43 & \bf79.21
\\
\hline
\end{tabular}

\caption{Unlabeled attachment scores (left) and labeled attachment scores (right) on the \textbf{test} sets. In each table, the first two columns show the results of the parser with word lookup (\textbf{Words}) vs.~character-based (\textbf{Chars}) representations. The last two columns add POS tags.  Boldface shows the better result comparing \textbf{Words} vs.~\textbf{Chars} and comparing \textbf{Words + POS} vs.~\textbf{Chars + POS}.\label{testresults} }
\end{table*}